\documentclass[runningheads]{llncs}
\usepackage{graphicx}
\usepackage{amsmath,amssymb} 
\usepackage{color}
\usepackage{multirow}
\usepackage{booktabs}
\usepackage[width=122mm,left=12mm,paperwidth=146mm,height=193mm,top=12mm,paperheight=217mm]{geometry}
\usepackage{subcaption}
\usepackage[colorlinks]{hyperref}

\begin{document}
\pagestyle{headings}
\mainmatter
\def\ECCV18SubNumber{2126}  

\title{Attend and Rectify: a Gated Attention Mechanism for Fine-Grained Recovery} 

\titlerunning{Attend and Rectify}

\authorrunning{Rodr\'iguez et al.}

\author{Pau Rodr\'{i}guez$^\dagger$, Josep M. Gonfaus$^\ddagger$, Guillem Cucurull$^\dagger$, \\ F. Xavier Roca$^{\dagger}$, Jordi Gonz\`{a}lez$^{\dagger}$}

\institute{$^\dagger$Computer Vision Center and Universitat Aut\`{o}noma de Barcelona (UAB), \\ Campus UAB, 08193 Bellaterra, Catalonia Spain \\
$^\ddagger$Visual Tagging Services, Parc de Recerca, Campus UAB}

\maketitle

\begin{abstract}
We propose a novel attention mechanism to enhance Convolutional
Neural Networks for fine-grained recognition. It learns to attend to lower-level feature activations without requiring part annotations and uses these activations to update and rectify the output likelihood distribution. In contrast to other approaches, the proposed mechanism is modular, architecture-independent and efficient both in terms of parameters and computation required. Experiments show that networks augmented with our approach systematically improve their classification accuracy and become more robust to clutter. As a result, Wide Residual Networks augmented with our proposal surpasses the state of the art classification accuracies in CIFAR-10, the Adience gender recognition task, Stanford dogs, and UEC Food-100.

\keywords{Deep Learning, Convolutional Neural Networks, Attention}
\end{abstract}

\section{Introduction}
Humans and animals process vasts amounts of information with limited computational resources thanks to attention mechanisms which allow them to focus resources on the most informative chunks of information \cite{anderson1985cognitive,desimone1995neural,ungerleider2000mechanisms} 

This work is inspired by the advantages of visual and biological attention mechanisms, for tackling fine-grained visual recognition with Convolutional Neural Networks (CNN) \cite{lecun1998gradient}. This is a particularly difficult task since it involves looking for details in large amounts of data (images) while remaining robust to deformation and clutter. In this sense, different attention mechanisms for fine-grained recognition exist in the literature: (i) iterative methods that process images using "glimpses" with recurrent neural networks (RNN) or long short-term memory (LSTM) \cite{sermanet2014attention,zhao2017diversified}, (ii) feed-forward attention mechanisms that augment vanilla CNNs, such as the Spatial Transformer Networks (STN) \cite{jaderberg2015spatial}, or top-down feed-forward attention mechanisms (FAM) \cite{rodriguez2017age}. Although it is not applied to fine-grained recognition, the Residual Attention introduced by \cite{wang2017residual} is another example of feed-forward attention mechanism that takes advantage of residual connections \cite{he2016deep} to enhance or dampen certain regions of the feature maps in an incremental manner.

\begin{figure}[t!]
\centering 
	\includegraphics[width=\textwidth]{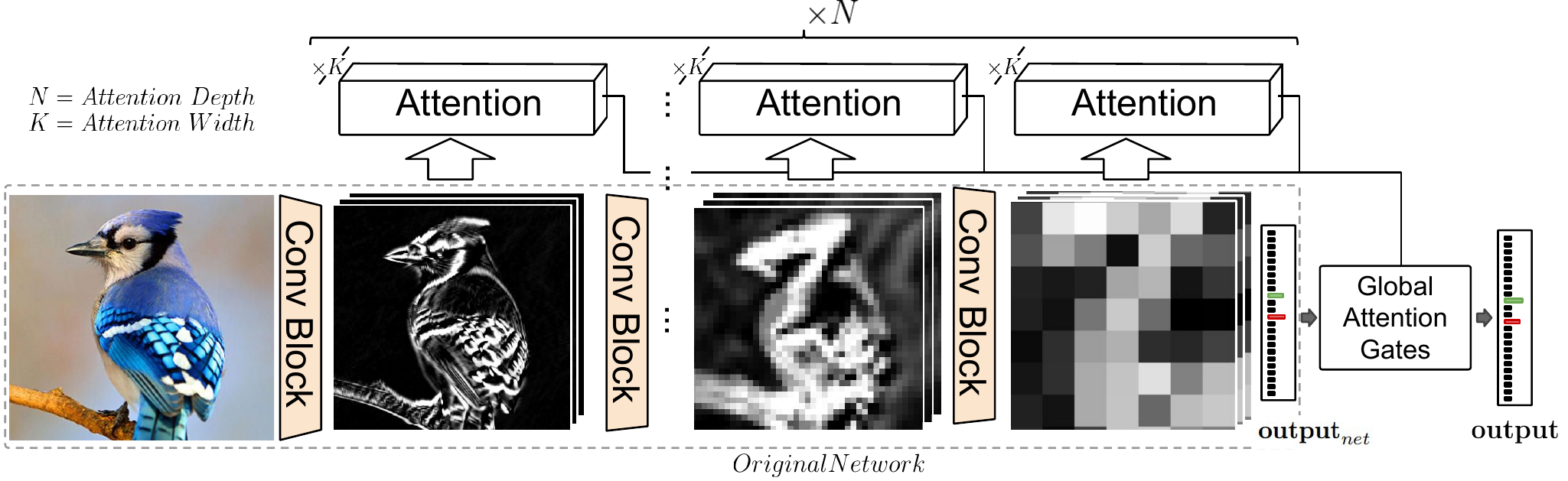}
\caption{The proposed mechanism. The original CNN is augmented with $N$ attention modules at $N$ different depths. Each attention module applies $K$ attention heads to the network feature maps to make a class prediction based on local information. The original network $\mathbf{output}_{net}$ is then corrected based on the local features by means of the global attention gates, resulting in the final $\mathbf{output}$. }
\label{fig:overall}
\end{figure}

Thus, most of the existing attention mechanisms are either limited by having to perform multiple passes through the data \cite{sermanet2014attention}, by carefully designed architectures that should be trained from scratch \cite{jaderberg2015spatial}, or by considerably increasing the needed amount of memory and computation, thus introducing computational bottlenecks \cite{jetley2018learn}. Hence, there is still the need of models with the following learning properties: (i) Detect and process in detail the most informative parts of an image for learning models more robust to deformation and clutter \cite{mnih2014recurrent}; (ii) feed-forward trainable with SGD for achieving faster inference than iterative models \cite{sermanet2014attention,zhao2017diversified}, together with faster convergence rate than Reinforcement Learning-based (RL) methods \cite{sermanet2014attention,liu2016fully}; (iii) preserve low-level detail for a direct access to local low-level features before they are modified by residual identity mappings. This is important for fine-grained recognition, where low-level patterns such as textures can help to distinguish two similar classes. This is not fulfilled by Residual Attention, where low-level features are subject to noise after traversing multiple residual connections \cite{wang2017residual}. 

In addition, desirable properties for attention mechanisms applied to CNNs would be: (i) \textbf{Modular and incremental}, since the same structure can be applied at each layer on any convolutional architecture, and it is easy to adapt to the task at hand; (ii) \textbf{Architecture independent}, that is, being able to adapt any pre-trained architecture such as VGG \cite{simonyan2014very} or ResNet \cite{he2016deep}; (iii) \textbf{Low computational impact} implying that it does not result in a significant increase in memory and computation; and (iv) \textbf{Simple} in the sense that it can be implemented in few lines of code, making it appealing to be used in future work.

Based on all these properties, we propose a novel attention mechanism that learns to attend low-level features from a standard CNN architecture through a set of replicable Attention Modules and gating mechanisms (see Section \ref{sect:approach}). Concretely, as it can be seen in Figure \ref{fig:overall}, any existing architecture can be augmented by applying the proposed model at different depths, and replacing the original loss by the proposed one. It is remarkable that the modules are independent of the original path of the network, so in practice, it can be computed in parallel to the rest of the network. The proposed attention mechanism has been included in a strong baseline like Wide Residual Networks (WRN) \cite{Zagoruyko2016WRN}, and applied on CIFAR-10, CIFAR-100 \cite{krizhevsky2009learning}, and five challenging fine-grained recognition datasets. The resulting network, called Wide Attentional Residual Network (WARN) systematically enhances the performance of WRNs and surpasses the state of the art in various classification benchmarks.

\section{Related Work}
There are different approaches to fine-grained recognition \cite{zhao2017survey}: (i) vanilla deep CNNs, (ii) CNNs as feature extractors for localizing parts and do alignment, (iii) ensembles, (iv) attention mechanisms. In this work, we focus on (iv), the attention mechanisms, which aim to discover the most discriminative parts of an image to be processed in greater detail, thus ignoring clutter and focusing on the most distinctive traits. These parts are central for fine-grained recognition, where the inter-class variance is small and the intra-class variance is high.

Different fine-grained attention mechanisms can be found in the literature. \cite{xiao2015application} proposed a \emph{two-level attention} mechanism for fine-grained classification on different subsets of the ILSVRC \cite{russakovsky2012imagenet} dataset, and the CUB200\_2011. In this model, images are first processed by a bottom-up object proposal network based on R-CNN \cite{zhang2014part} and selective search \cite{uijlings2013selective}. Then, the softmax scores of another ILSVRC2012 pre-trained CNN, which they call \emph{FilterNet}, are thresholded to prune the patches with the lowest parent class score. These patches are then classified to fine-grained categories with a \emph{DomainNet}. Spectral clustering is also used on the \emph{DomainNet} filters in order to extract parts (head, neck, body, etc.), which are classified with an SVM. Finally, the part- and object-based classifier scores are merged to get the final prediction. The \emph{two-level attention} obtained state of the art results on CUB200-2011 with only class-level supervision. However, the pipeline must be carefully fine-tuned since many stages are involved with many hyper-parameters.

Differently from \emph{two-level attention}, which consists of independent processing and it is not end-to-end, Sermanet \emph{et al.} proposed to use a deep CNN and a Recurrent Neural Network (RNN) to accumulate high multi-resolution ``glimpses'' of an image to make a final prediction \cite{sermanet2014attention}. However, reinforcement learning slows down convergence and the RNN adds extra computation steps and parameters.

A more efficient approach was presented by Liu \emph{et al.} \cite{liu2016fully}, where a fully-convolutional network is trained with reinforcement learning to generate confidence maps on the image and use them to extract the parts for the final classifiers whose scores are averaged. Compared to previous approaches, in the work done by \cite{liu2016fully}, multiple image regions are proposed in a single timestep thus, speeding up the computation. A greedy reward strategy is also proposed in order to increase the training speed. The recent approach presented by \cite{fu2017look} uses a classification network and a recurrent attention proposal network that iteratively refines the center and scale of the input (RA-CNN). A ranking loss is used to enforce incremental performance at each iteration.

Zhao \emph{et al.} proposed to enforce multiple non-overlapped attention regions \cite{zhao2017diversified}. The overall architecture consists of an attention canvas generator, which extracts patches of different regions and scales from the original image; a VGG-16 \cite{simonyan2014very} CNN is then used to extract features from the patches, which are aggregated with a long short-term memory \cite{hochreiter1997long} that attends to non-overlapping regions of the patches. Classification is performed with the average prediction in each region. Similarly, in \cite{zheng2017learning}, they proposed the Multi-Attention CNN (MA-CNN) to learn to localize informative patches from the output of a VGG-19 and use them to train an ensemble of part classifiers.

In \cite{jetley2018learn}, they propose to extract global features from the last layers of a CNN, just before the classifier and use them to attend relevant regions in lower level feature activations. The attended activations from each level are then spatially averaged, channel-wise concatenated, and fed to the final classifier. The main differences with \cite{jetley2018learn} are: (i) attention maps are computed in parallel to the base model, while the model in \cite{jetley2018learn} requires output features for computing attention maps; (ii) WARN uses fewer parameters, so dropout is not needed to obtain competitive performance (these two factors clearly reflect in gain of speed); and (iii) gates allow our model to ignore/attend different information to improve the performance of the original model, while in \cite{jetley2018learn} the full output function is replaced. As a result, WARN obtains 3.44\% error on CIFAR10, outperforming \cite{jetley2018learn} while being 7 times faster w/o parallelization. 

All the previously described methods involve multi-stage pipelines and most of them are trained using reinforcement learning (which requires sampling and makes them slow to train). In contrast, STNs, FAM, the model in \cite{jetley2018learn}, and our approach jointly propose the attention regions and classify them in a single pass. Moreover, different from STNs and FAM our approach only uses one CNN stream, it can be used on pre-trained models, and it is far more computationally efficient than STNs, FAM, and \cite{jetley2018learn} as described next.

\section{Our approach}
\label{sect:approach}

\begin{figure}[t!]
\centering 
\begin{subfigure}{0.55\textwidth}
	\includegraphics[width=\textwidth]{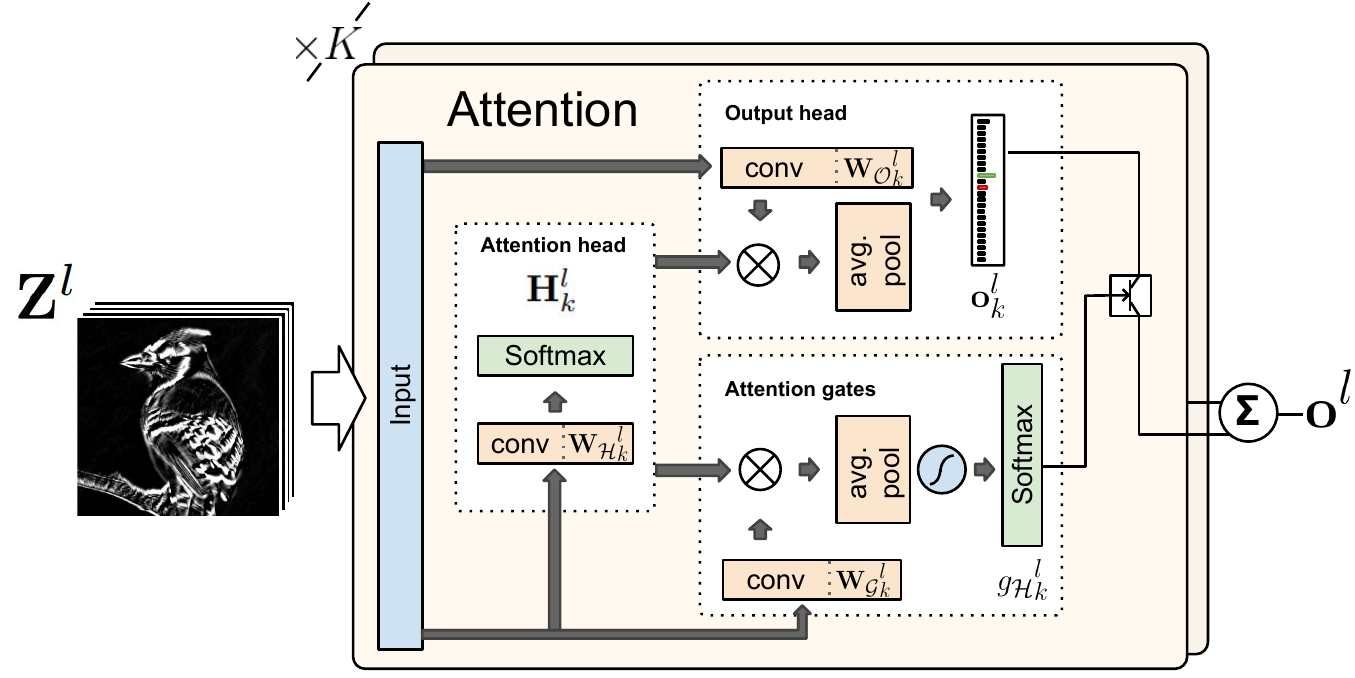}
\caption{Attention module}
\label{fig:attention_module}
\end{subfigure}
\begin{subfigure}{0.4\textwidth}
	\includegraphics[width=\textwidth]{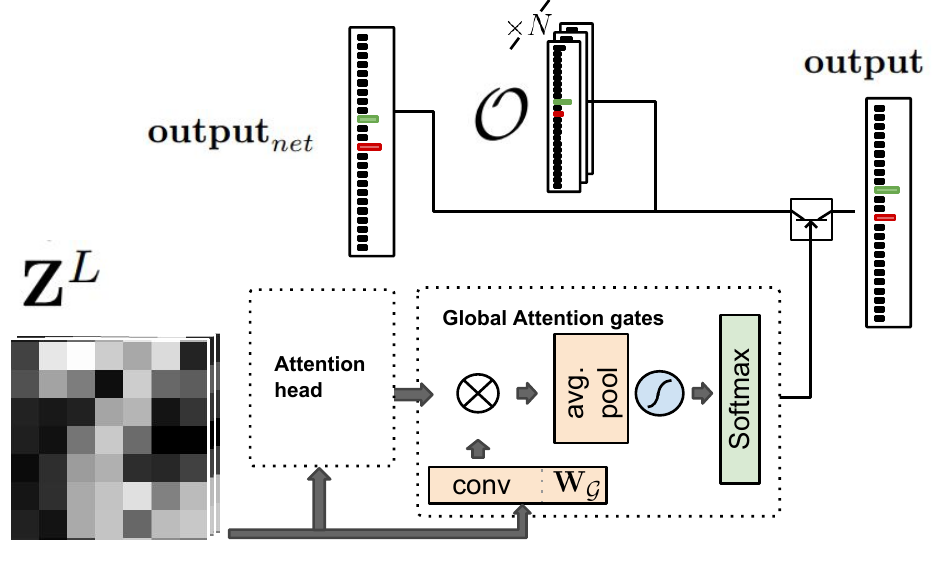}
\caption{Attention gates}
\label{fig:attention_gates}
\end{subfigure}
\caption{(a) Attention Module: $K$ attention heads $\mathbf{H}^l_k$ are applied to a feature map $\mathbf{Z}^l$, and information is aggregated with the layer attention gates. (b) Global attention: global information from the last feature map $\mathbf{Z}^L$ is used to compute the gating scores that produce the final $\mathbf{output}$ as the weighted average of the outputs of the attention modules and the original network $\mathbf{output}_{net}$}
\label{fig:modules}
\end{figure}

Our approach consists of a universal attention module that can be added after each convolutional layer without altering pre-defined information pathways of any architecture (see Figure \ref{fig:overall}). This is helpful since it seamlessly augments any architecture such as VGG and ResNet with no extra supervision, \emph{i.e.} no part labels are necessary. Furthermore, it also allows being plugged into any existing trained network to quickly perform transfer learning approaches.

The attention module consists of three main submodules depicted in Figure \ref{fig:modules} (a): (i) the attention heads $\mathcal{H}$, which define the most relevant regions of a feature map, (ii) the output heads $\mathcal{O}$, generate an hypothesis given the attended information, and (iii) the confidence gates $\mathcal{G}$, which output a confidence score for each attention head. Each of these modules is described in detail in the following subsections.

\subsection{Overview}
As it can be seen in Figure \ref{fig:overall}, a convolution layer is applied to the output of the augmented layer, producing $K$ attentional heatmaps. These attentional maps are then used to spatially average the local class probability scores for each of the feature maps, and produce the final class probability vector. This process is applied to an arbitrary number $N$ of layers, producing $N$ class probability vectors. Then, the model learns to correct the initial prediction by attending the lower-level class predictions. This is the final combined prediction of the network. In terms of probability, the network corrects the initial likelihood by updating the prior with local information.

\subsection{Attention head}
Inspired by \cite{zhao2017diversified} and the \emph{transformer} architecture presented by \cite{vaswani2017attention}, and following the notation established by \cite{Zagoruyko2016WRN}, we have identified two main dimensions to define attentional mechanisms: (i) the number of layers using the attention mechanism, which we call \emph{attention depth} (AD), and (ii) the number of attention heads in each attention module, which we call \emph{attention width} (AW). Thus, a desirable property for any universal attention mechanism is to be able to be deployed at any arbitrary \emph{depth} and \emph{width}. 

This property is fulfilled by including $K$ attention heads $\mathcal{H}_k$ (width), depicted in Figure \ref{fig:overall}, into each attention module (depth)\footnote{Notation: $\mathcal{H,O,G}$ are the set of attention heads, output heads, and attention gates respectively. Uppercase letters refer to functions or constants, and lowercase ones to indices. Bold uppercase letters represent matrices and bold lowercase ones vectors.}. Then, the attention heads at layer $l\in [1..L]$, receive the feature activations $\mathbf{Z}^l \in \mathbb{R}^{c \times h\times w}$ of that layer as input, and output $K$ attention masks:

\begin{equation}
\mathbf{H}^l = spatial\_softmax (\mathbf{W_\mathcal{H}}^l \ast \mathbf{Z}^l),
\label{eq:atthead1}
\end{equation}

where $\mathbf{H}^l \in \mathbb{R}^{K\times h\times w}$ is the output matrix of the $l^{th}$ attention module, $\mathbf{W_\mathcal{H}}^l: \mathbb{R}^{c \times h \times w} \rightarrow \mathbb{R}^{K \times h \times w}$ is a convolution kernel with output dimensionality $K$ used to compute the attention masks corresponding to the attention heads $\mathbf{H}_k$, and $\ast$ denotes the convolution operator. The \emph{spatial\_softmax}, which performs the \emph{softmax} operation channel-wise on the spatial dimensions of the input, is used to enforce the model to learn the most relevant region of the image. Sigmoid units could also be used at the risk of degeneration to all-zeros or all-ones. To prevent the attention heads at the same depth to collapse into the same region, we apply the regularizer proposed in \cite{zhao2017diversified}.

\subsection{Output head}
To obtain the class probability scores, the input feature map   $\mathbf{Z}^l_k$ is convolved with a kernel:

$$
\mathbf{W_\mathcal{O}}^l_k \in \mathbb{R}^{channels \times h\times w} \rightarrow \mathbb{R}^{\#labels \times h \times w},
$$

$h,w$ represent the spatial dimensions, and $channels$ is the number of input channels to the module. This results on a spatial map of class probability scores:

\begin{equation}
\mathbf{O}^l_k = \mathbf{W_\mathcal{O}}^l_k \ast \mathbf{Z}^l.
\end{equation}

Note that this operation can be done in a single pass for all the $K$ heads by setting the number of output channels to $\#labels \cdot K$. Then, class probability vectors $\mathbf{O}_k^l$ are weighted by the spatial attention scores and spatially averaged:

\begin{equation}
\mathbf{o}^l_k = \sum_{x,y} \mathbf{H}^l_k \odot \mathbf{O}^l_k,
\label{eq:atthead2}
\end{equation}

where $\odot$ is the element-wise product, and $x\in \{1..width\}, y\in\{1..height\}$. The attention scores $\mathbf{H}^l_k$ are a 2d flat mask and the product with each of the input channels of $\mathbf{Z^l}$ is done by broadcasting, \emph{i.e.} repeating $\mathbf{H}^l_k$ for each of the channels  of $\mathbf{Z^l}$.

\subsection{Layered attention gates}

The final output $\mathbf{o}^l$ of an attention module is obtained by a weighted average of the $K$ output probability vectors, through the use of head attention gates ${\mathbf{g}_\mathcal{H}}^l \in \mathbb{R}^{|\mathcal{H}|},\ \sum_k {g_\mathcal{H}}^l_k = 1$.

\begin{equation}
\mathbf{o}^l = \sum_k {g_\mathcal{H}}^l_k o^l_k.
\end{equation}

Where $g_\mathcal{H}$ is obtained by first convolving $\mathbf{Z}^l$ with

$$
\mathbf{W_g}^l \in \mathbb{R}^{channels \times h\times w} \rightarrow \mathbb{R}^{|\mathcal{H}| \times h \times w},
$$

and then performing a spatial weighted average:

\begin{equation}
\mathbf{g_\mathcal{H}}^l = softmax(tanh(\sum_{x,y} (\mathbf{W_g}^l \ast \mathbf{Z}^l ) \odot \mathbf{H}_l)).
\end{equation}

This way, the model learns to choose the attention head that provides the most meaningful output for a given attention module.

\subsection{Global attention gates}
In order to let the model learn to choose the most discriminative features at each depth to disambiguate the output prediction, a set of relevance scores $\mathbf{c}$  are predicted at the model output, one for each attention module, and one for the final prediction. This way, through a series of gates, the model can learn to query information from each level of the network conditioned to the global context. Note that, unlike in \cite{jetley2018learn}, the final predictions do not act as a bottleneck to compute the output of the attention modules.

The relevance scores are obtained with an inner product between the last feature activation of the network $\mathbf{Z}^L$ and the gate weight matrix $\mathbf{W}_\mathcal{G}$:

\begin{equation}
\label{eq:scores}
\mathbf{c} = tanh(\mathbf{W_\mathcal{G}}\mathbf{Z}^L).
\end{equation}

The gate values $\mathbf{g_\mathcal{O}}$ are then obtained by normalizing the set of scores by means of a $softmax$ function:

\begin{equation}
\label{eq:gates}
{g_\mathcal{O}}^l_k = \frac{e^{c^l_k}}{\sum_{i = 1}^{|\mathcal{G}|} e^{c_i}},
\end{equation}

where $|\mathcal{G}|$ is the total number of gates, and $c_i$ is the $i^{th}$ confidence score from the set of all confidence scores. The final output of the network is the weighted sum of the attention modules: 

\begin{equation}
\label{eq:all}
\mathbf{output} = g_{net} \cdot \mathbf{output}_{net} + \sum_{l \in \{1..|\mathcal{O}|\}} g_\mathcal{O}^l \cdot \mathbf{o}^l,
\end{equation}

where $g_{net}$ is the gate value for the original network output ($\mathbf{output_{net}}$), and $\mathbf{output}$ is the final output taking the attentional predictions $\mathbf{o}^l$ into consideration. Note that setting the output of $\mathcal{G}$ to $\frac{1}{|\mathcal{O}|}$, corresponds to averaging all the outputs. Likewise, setting $\{\mathcal{G} \setminus G_{output}\} = 0, G_{output} = 1$, \emph{i.e.} the set of attention gates is set to zero and the output gate to one, corresponds to the original pre-trained model without attention.

It is worth noting that all the operations that use $\mathbf{Z}^l$ can be aggregated into a single convolution operation. Likewise, the fact that the attention mask is generated by just one convolution operation, and that most masking operations are directly performed in the label space, or can be projected into a smaller dimensionality space, makes the implementation highly efficient. Additionally, the direct access to the output gradients makes the module fast to learn, thus being able to generate foreground masks from the beginning of the training and refining them during the following epochs. 

\section{Experiments}
We empirically demonstrate the impact on the accuracy and robustness of the different modules in our model on Cluttered Translated MNIST and then compare it with state-of-the-art models such as DenseNets and ResNeXt. Finally, we demonstrate the universality of our method for fine-grained recognition through a set of experiments on five fine-grained recognition datasets, as detailed next.

\subsection{Datasets}
\begin{figure}[t!]
\centering
\begin{subfigure}[t]{0.19\textwidth}
\centering
\includegraphics[width=0.8\textwidth]{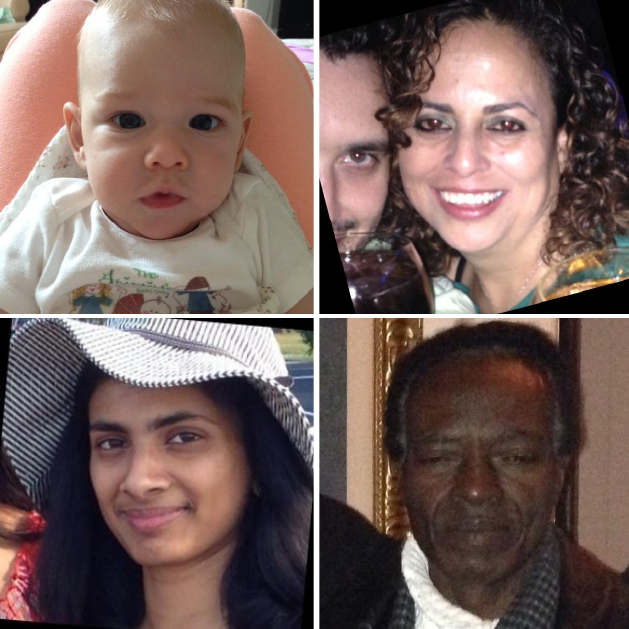}
\caption{}
\label{fig:adience_sample}
\end{subfigure}
\begin{subfigure}[t]{0.19\textwidth}
\centering
\includegraphics[width=0.8\textwidth]{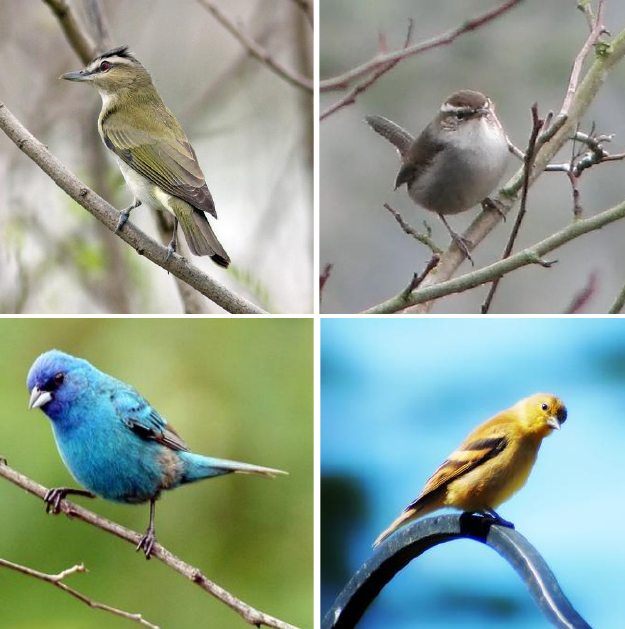}
\caption{}
\label{fig:birds_sample}
\end{subfigure}
\begin{subfigure}[t]{0.19\textwidth}
\centering
\includegraphics[width=0.8\textwidth]{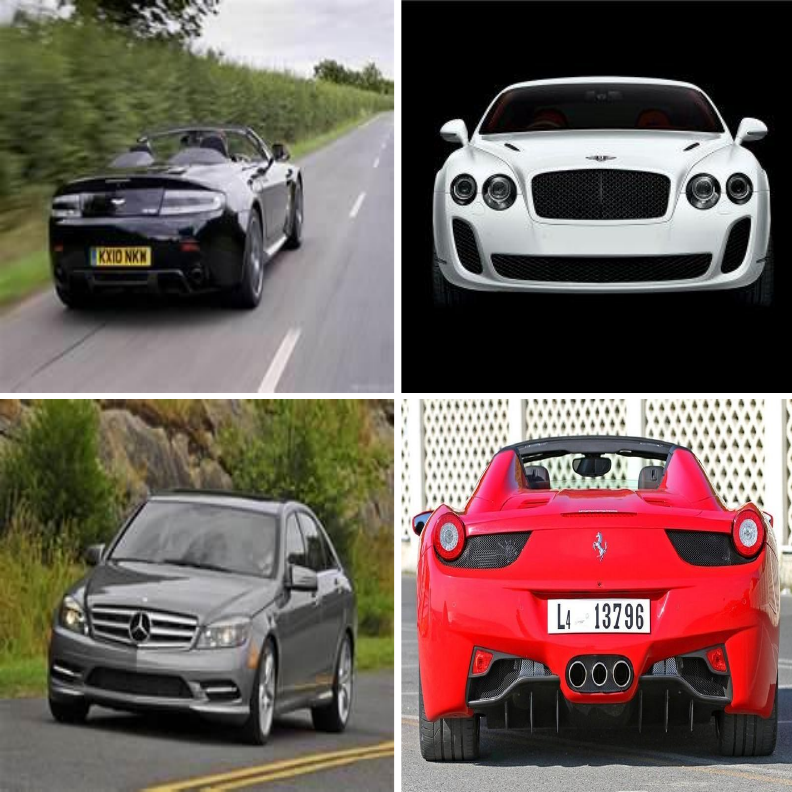}
\caption{}
\label{fig:cars_sample}
\end{subfigure}
\begin{subfigure}[t]{0.19\textwidth}
\centering
\includegraphics[width=0.8\textwidth]{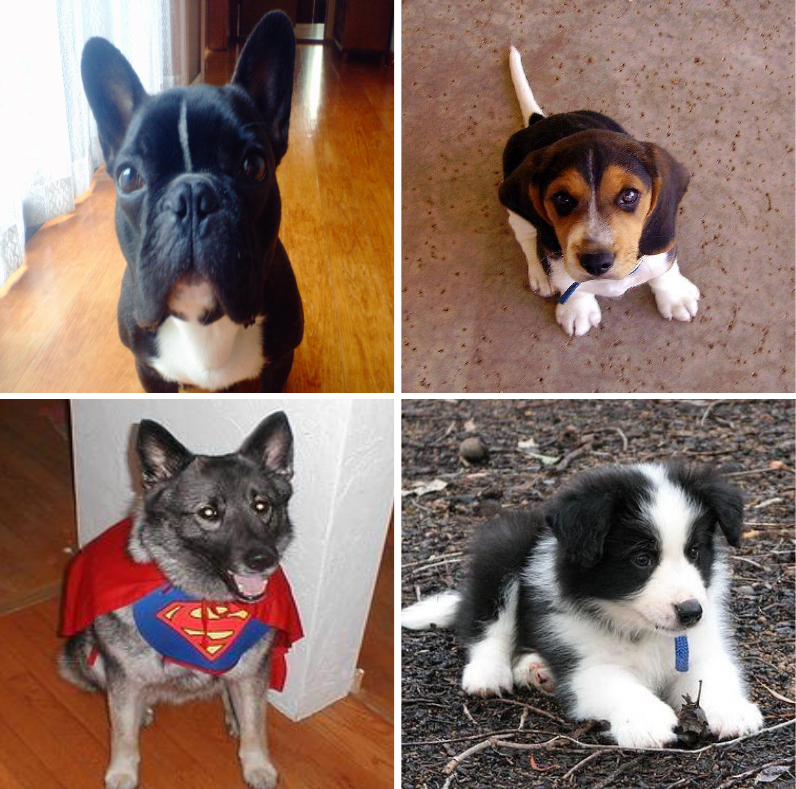}
\caption{}
\label{fig:dogs_sample}
\end{subfigure}
\begin{subfigure}[t]{0.19\textwidth}
\centering
\includegraphics[width=0.8\textwidth]{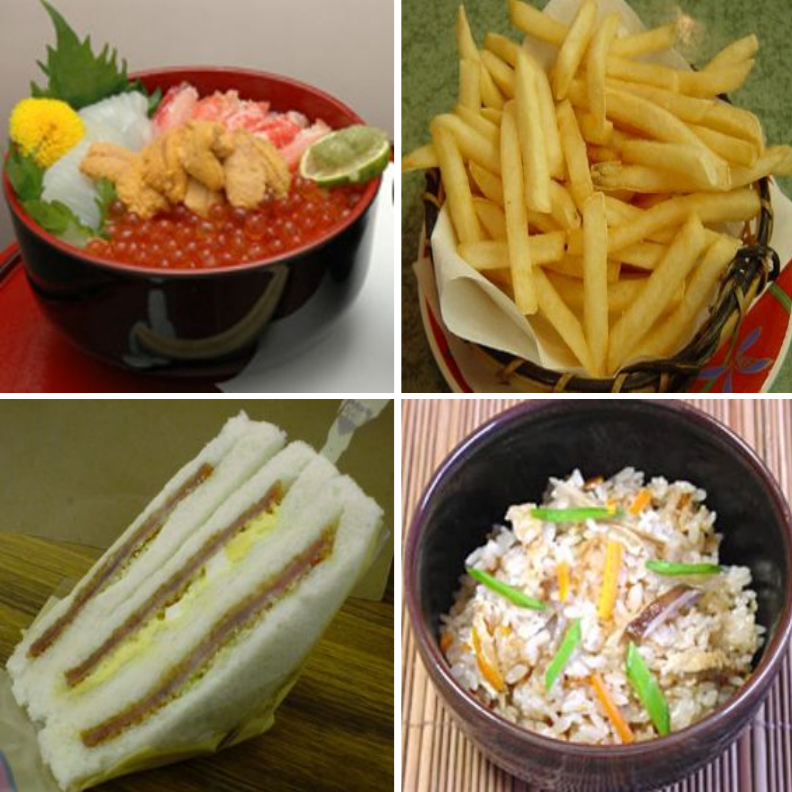}
\caption{}
\label{fig:food_sample}
\end{subfigure}
\caption{Samples from the five fine-grained datasets. (a) Adience, (b) CUB200 Birds, (c) Stanford Cars, (d) Stanford Dogs, (e) UEC-Food100}
\end{figure}

\noindent\textbf{Cluttered Translated MNIST}\footnote{\url{https://github.com/deepmind/mnist-cluttered}}
Consists of $40\times 40$ images containing a randomly placed MNIST \cite{lecun1998mnist} digit and a set of $D$ randomly placed distractors, see Figure \ref{fig:clutter_mnist}. The distractors are random $8\times 8$ patches from other MNIST digits.

\medskip
\noindent\textbf{CIFAR}\footnote{\url{https://www.cs.toronto.edu/~kriz/cifar.html}}
The CIFAR dataset consists of 60K 32x32 images in 10 classes for CIFAR-10, and 100 for CIFAR-100. There are 50K training and 10K test images. 

\medskip
\noindent\textbf{Stanford Dogs \cite{khosla2011novel}.}
The Stanford Dogs dataset consists of 20.5K images of 120 breeds of dogs, see Figure \ref{fig:dogs_sample}. The dataset splits are fixed and they consist of 12k training images and 8.5K validation images. 

\medskip
\noindent\textbf{UEC Food 100 \cite{matsuda12}.} A Japanese food dataset with 14K images of 100 different dishes, see Figure \ref{fig:food_sample}. In order to follow the standard procedure (\emph{e.g.}  \cite{chen2016deep,hassannejad2016food}), bounding boxes are used to crop the images before training.

\medskip
\noindent\textbf{Adience dataset \cite{eidinger2014age}.} The adience dataset consists of 26.5 K images distributed in eight age categories (0–2, 4–6, 8–13, 15–20, 25–32, 38–43, 48–53, 60+), and gender labels. A sample is shown in Figure \ref{fig:adience_sample}.
The performance on this dataset is measured using 5-fold cross-validation.

\medskip
\noindent\textbf{Stanford Cars \cite{krause20133d}.}
The Cars dataset contains 16K images of 196 classes of cars, see Figure \ref{fig:cars_sample}. The data is split into 8K training and 8K testing images.

\medskip
\noindent\textbf{Caltech-UCSD Birds 200 \cite{WahCUB_200_2011}.}
The CUB200-2011 birds dataset (see Figure \ref{fig:birds_sample}) consists of 6K train and 5.8K test bird images distributed in 200 categories. Although bounding box, segmentation, and attributes are provided, we perform raw classification as done by \cite{jaderberg2015spatial}. 

\subsection{Ablation study}
\label{sect:ablation}
\begin{figure}[!t]
\centering
\begin{subfigure}[!t]{0.26\textwidth}
\centering
\includegraphics[width=0.9\textwidth]{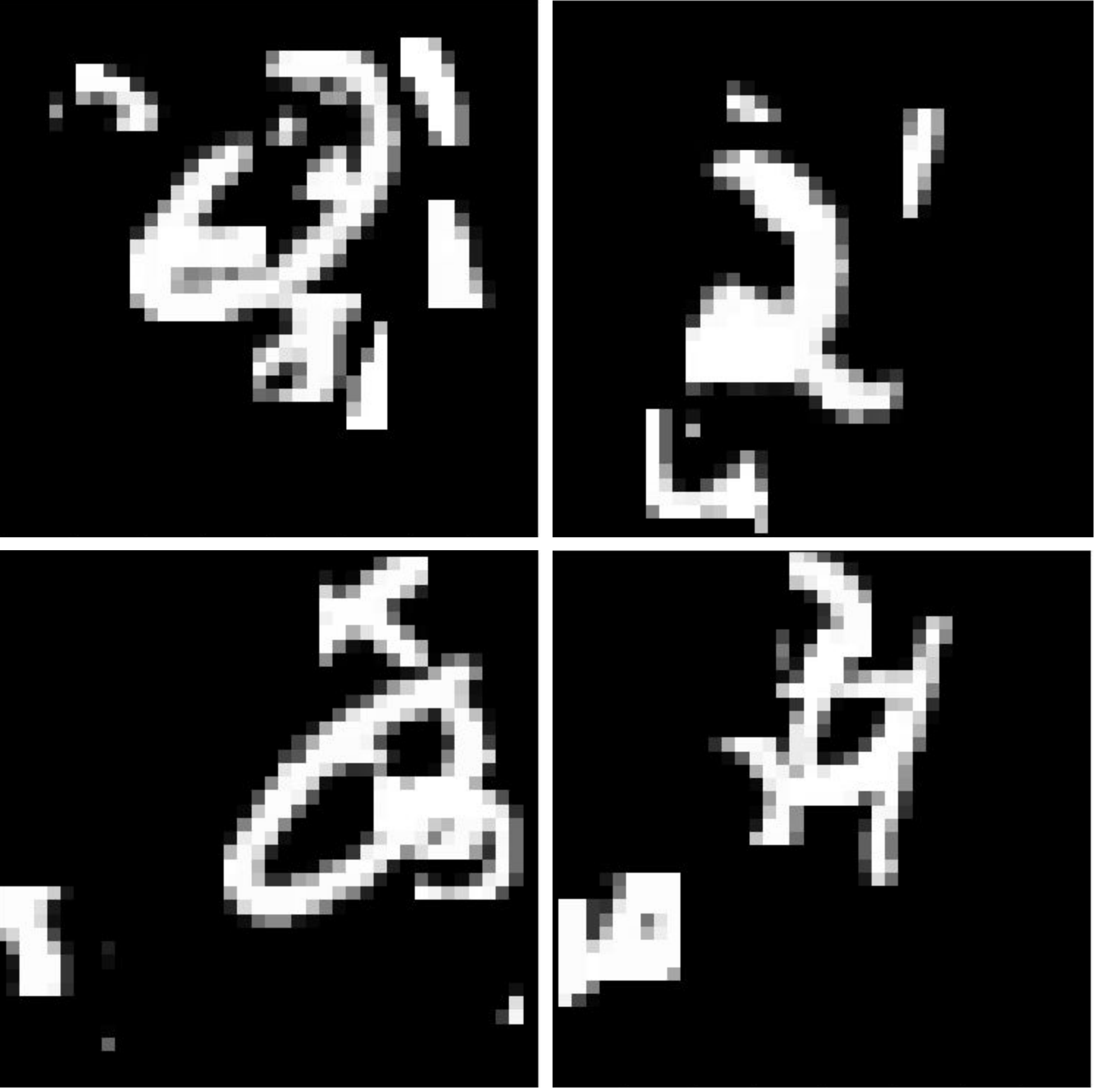}
\caption{Cluttered MNIST}
\label{fig:clutter_mnist}
\end{subfigure}
\begin{subfigure}[!t]{0.32\textwidth}
\centering
\includegraphics[width=\textwidth]{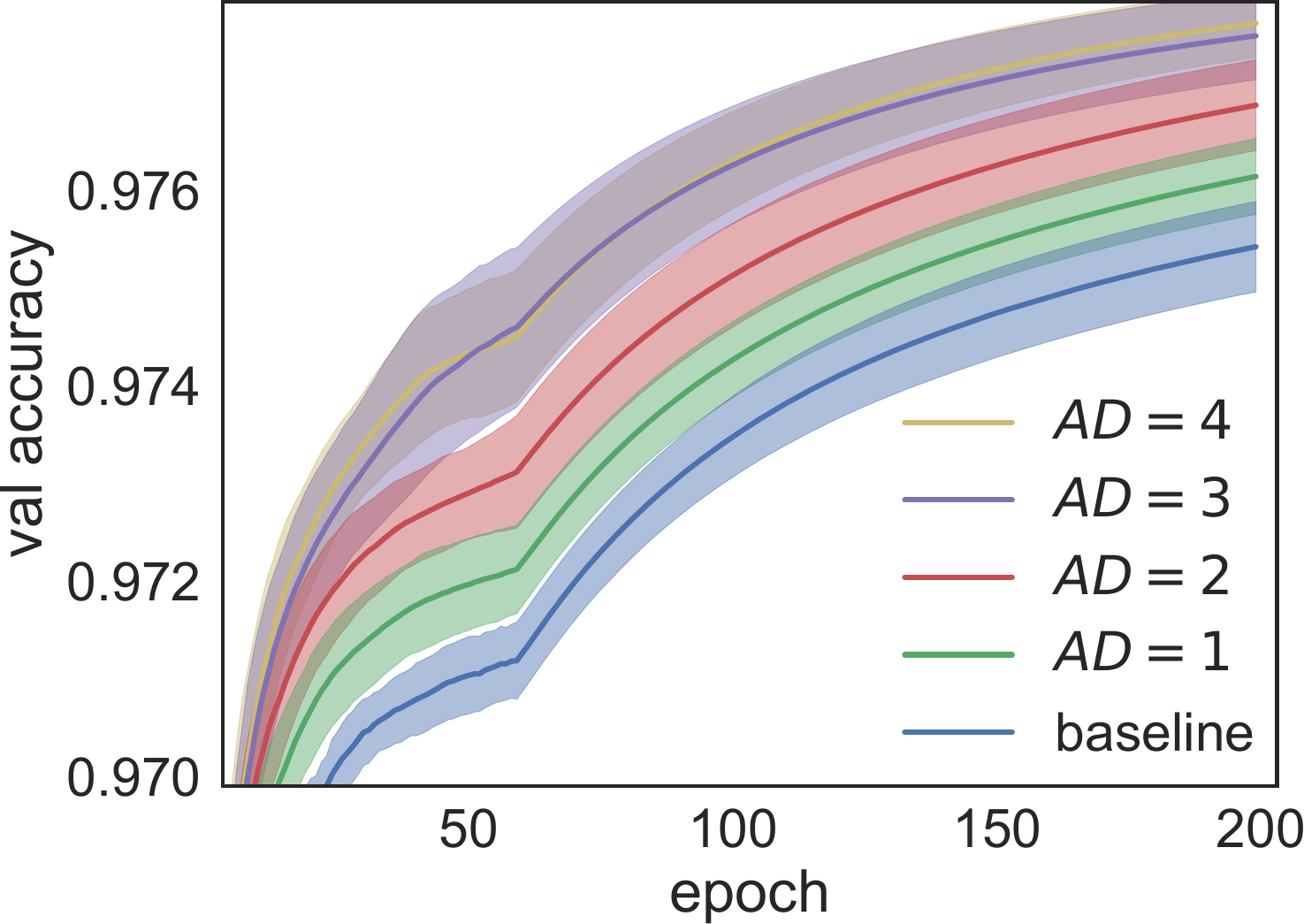}
\caption{Different depths}
\label{fig:ablation_depth}
\end{subfigure}
\begin{subfigure}[!t]{0.32\textwidth}
\centering
\includegraphics[width=\textwidth]{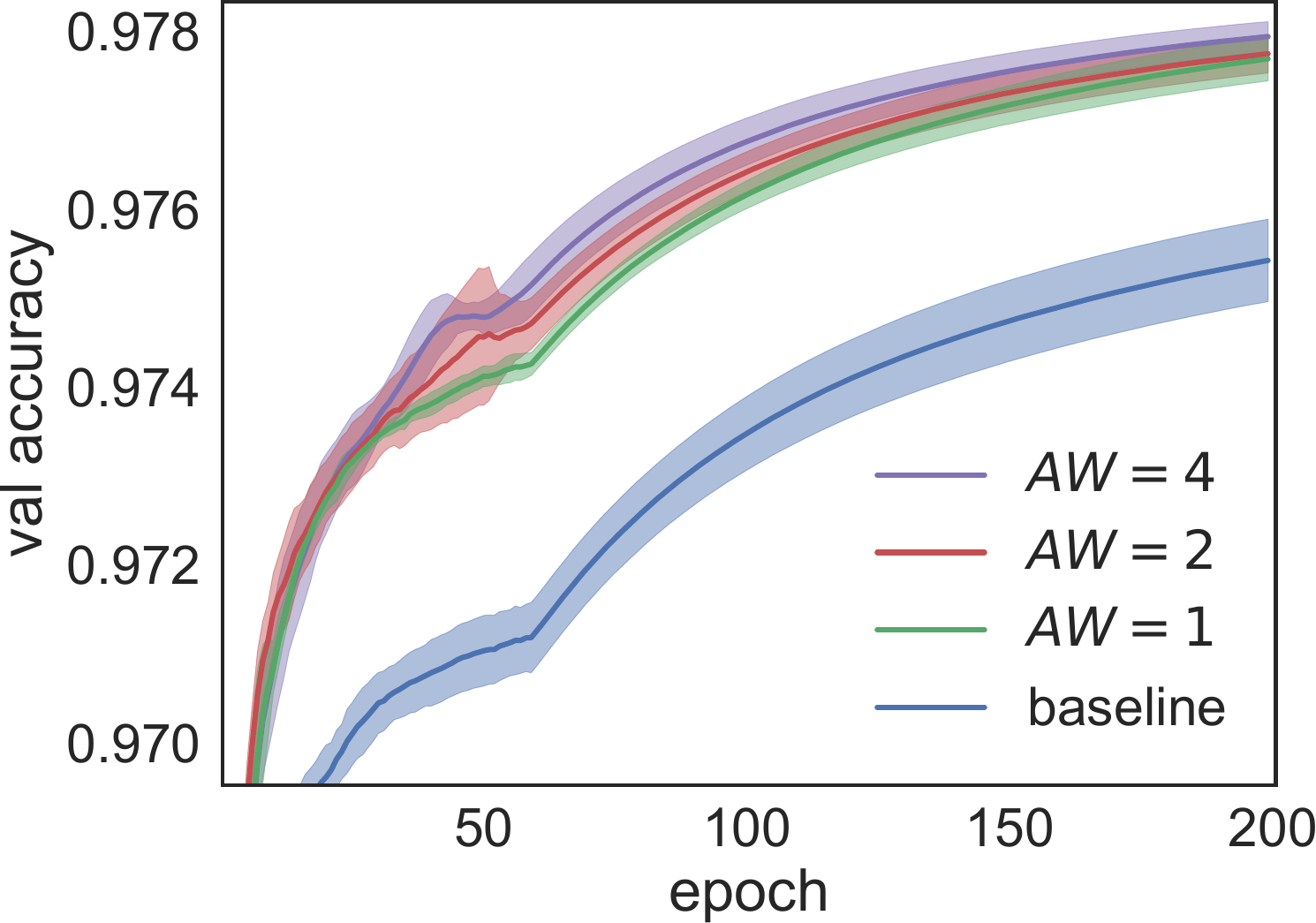}
\caption{Different widths}
\label{fig:ablation_width}
\end{subfigure}
\begin{subfigure}[!t]{0.32\textwidth}
\centering
\includegraphics[width=\textwidth]{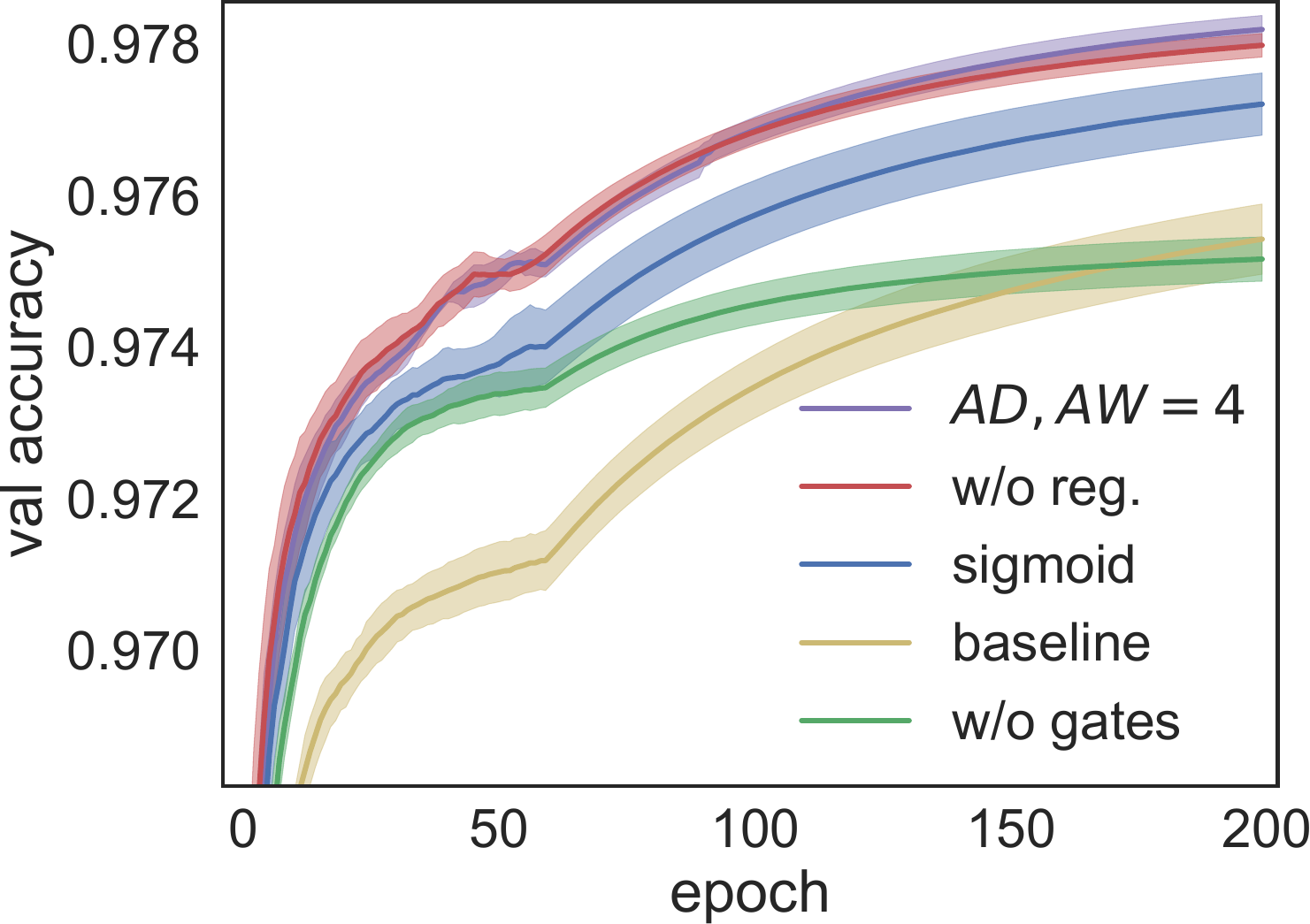}
\caption{Softmax, gates, reg.}
\label{fig:hyperparams}
\end{subfigure}
\begin{subfigure}[!t]{0.32\textwidth}
\centering
\includegraphics[width=\textwidth]{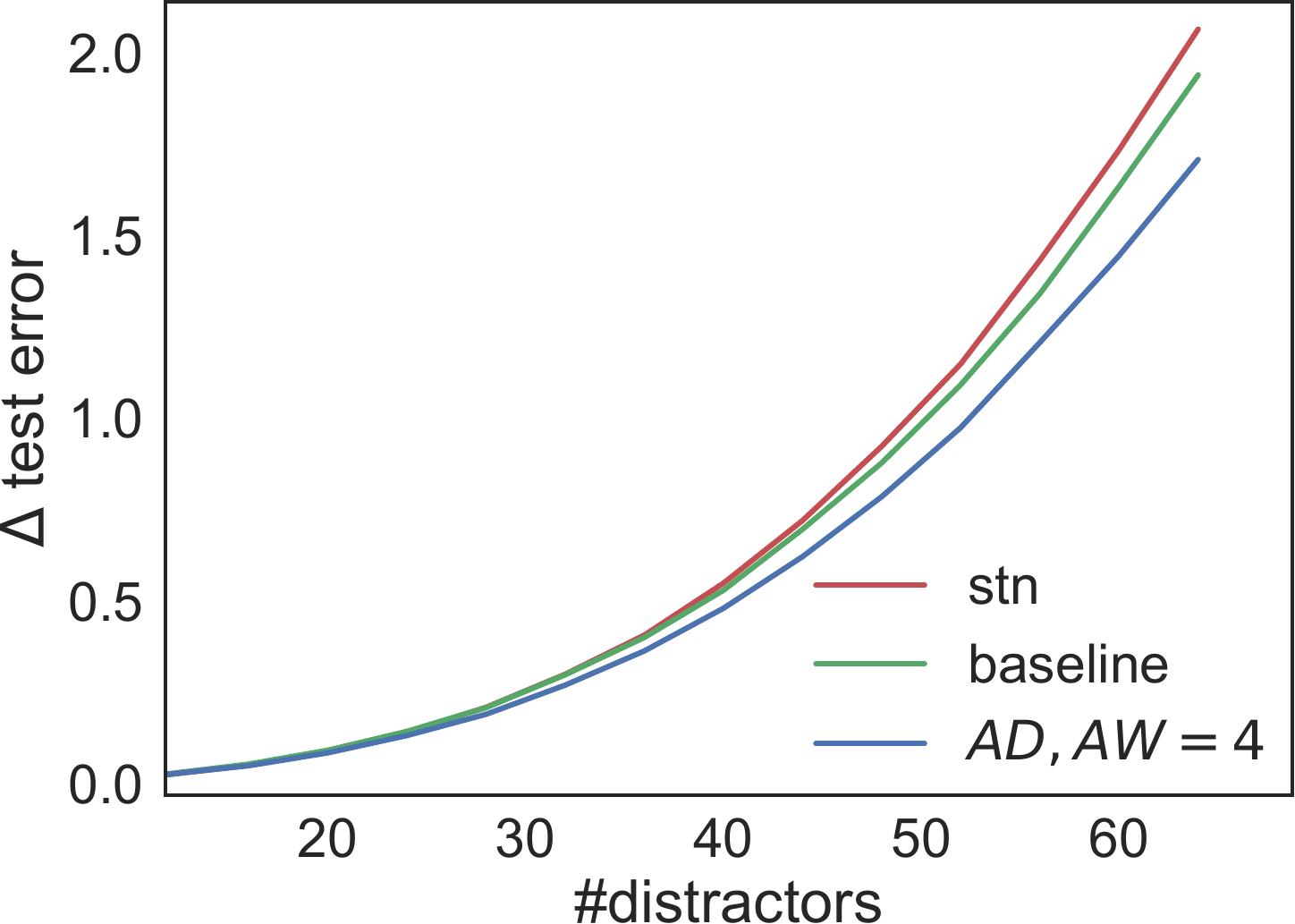}
\caption{Overfitting}
\label{fig:distractors}
\end{subfigure}
\caption{Ablation experiments on Cluttered Translated MNIST. \texttt{baseline} indicates the original model before being augmented with attention. (a) shows a sample of the cluttered MNIST dataset. (b) the effect of increasing the attention depth (AD), for attention width $AW=1$. (c) effect of increasing AW, for AD=4. (d) best performing model ($AD,AW=4$, softmax attention gates, and regularization \cite{zhao2017diversified}) vs unregularized, sigmoid attention, and without gates. (e) test error of the baseline, attention ($AD,AW=4$), and spatial transformer networks (stn), when trained with different amounts of distractors. }.
\label{fig:ablation}
\end{figure}

We evaluate the submodules of our method on the Cluttered Translated MNIST following the same procedure as in \cite{mnih2014recurrent}. The proposed attention mechanism is used to augment a CNN with five $3\times 3$ convolutional layers and two fully-connected layers in the end. The three first convolution layers are followed by Batch-normalization and a spatial pooling. Attention modules are placed starting from the fifth convolution (or pooling instead) backward until $AD$ is reached. Training is performed with SGD for $200$ epochs, and a learning rate of $0.1$, which is divided by $10$ after epoch $60$. Models are trained on a $200k$ images train set, validated on a $100k$ images validation set, and tested on $100k$ test images. Weights are initialized using He \emph{et al.}  \cite{he2015delving}. Figure \ref{fig:ablation} shows the effects of the different hyperparameters of the proposed model. The performance without attention is labeled as \texttt{baseline}. Attention models are trained with softmax attention gates and regularized with \cite{zhao2017diversified}, unless explicitly specified.

First, we test the importance of $AD$ for our model by increasingly adding attention layers with $AW=1$ after each pooling layer. As it can be seen in Figure \ref{fig:ablation_depth}, greater $AD$ results in better accuracy, reaching saturation at $AD = 4$, note that for this value the receptive field of the attention module is $5\times5\ px$, and thus the performance improvement from such small regions is limited. Figure \ref{fig:ablation_width} shows training curves for different values of $AW$, and $AD=4$. As it can be seen, small performance increments are obtained by increasing the number of attention heads even with a single object present in the image. 

Then, we use the best $AD$ and $AW$, \emph{i.e.} $AD,AW=4$, to verify the importance of using softmax on the attention masks instead of sigmoid (\ref{eq:atthead1}), the effect of using gates (Eq. \ref{eq:gates}), and the benefits of regularization \cite{zhao2017diversified}. Figure \ref{fig:hyperparams} confirms that ordered by importance: gates, softmax, and regularization result in accuracy improvement, reaching $97.8\%$. In particular, gates play an important role in discarding the distractors, especially for high $AW$ and high $AD$

Finally, in order to verify that attention masks are not overfitting on the data, and thus generalize to any amount of clutter, we run our best model so far (Figure \ref{fig:hyperparams}) on the test set with an increasing number of distractors (from 4 to 64). For the comparison, we included the baseline model before applying our approach and the same baseline augmented with an STN \cite{jaderberg2015spatial} that reached comparable performance as our best model in the validation set. All three models were trained with the same dataset with eight distractors. Remarkably, as it can be seen in Figure \ref{fig:distractors}, the attention augmented model demonstrates better generalization than the baseline and the STN.

\subsection{Training from scratch}
We benchmark the proposed attention mechanism on CIFAR-10 and CIFAR-100, and compare it with the state of the art. As a base model, we choose Wide Residual Networks, a strong baseline with a large number of parameters so that the additional parameters introduced by our model (WARN) could be considered negligible. The same WRN baseline is used to train an \texttt{att2} model \cite{jetley2018learn}, we refer to this model as WRN-att2. Models are initialized and optimized following the same procedure as in \cite{Zagoruyko2016WRN}. Attention Modules are systematically placed after each of the three convolutional groups starting by the last one until the attention depth has been reached in order to capture information at different levels of abstraction and fine-grained resolution, this same procedure is followed in \cite{jetley2018learn}. The model is implemented with pytorch \cite{paszke2017pytorch} and run on a single workstation with two NVIDIA 1080Ti.\footnote{\url{https://github.com/prlz77/attend-and-rectify}}

First, the same ablation study performed in Section \ref{sect:ablation} is repeated on CIFAR100. We consistently reached the same conclusions as in Cluttered-MNIST: accuracy improves 1.5\% by increasing attention depth from 1 to \#residual\_blocks, and width from 1 to 4. Gating performs 4\% better than a simpler linear projection, and 3\% with respect to simply averaging the output vectors. A 0.6\% improvement is also observed when regularization is activated. Interestingly, we found sigmoid attention to perform similarly to softmax. With this setting, WARN reaches 17.82\% error on CIFAR100. In addition, we perform an experiment blocking the gradients from the proposed attention modules to the original network to analyze whether the observed improvement is due to the attention mechanism or an optimization effect due to introducing shortcut paths to the loss function \cite{lee2015deeply}. Interestingly, we observed a 0.2\% drop on CIFAR10, and 0.4\% on CIFAR100, which are still better than the baseline. Note that a performance drop should be expected, even without taking optimization into account, since backpropagation makes intermediate layers learn to gather more discriminative features for the attention layers. It is also worth noting that fine-grained accuracy improves even when fine-tuning (gradients are multiplied by 0.1 in the base model), see Section \ref{sect:transfer}. In contrast, the approach in \cite{jetley2018learn} does not converge when gradients are not sent to the base model since classification is directly performed on intermediate feature maps (which continuously shift during training).

\begin{table}[t!]
\centering
\caption{Error rate on CIFAR-10 and CIFAR-100 (\%). Results that surpass all other methods are in blue, results that surpass the baseline are in black bold font. Total network depth, attention depth, attention width, the usage of dropout, and the amount of floating point operations (Flop) are provided in columns 1-5 for fair comparison}
\label{tab:accuracy-benchmark}
\begin{tabular}{@{}lcrccccc@{}}
\toprule
 & \textbf{Net Depth} & \multicolumn{1}{c}{\textbf{AD}} & \textbf{AW} & \textbf{Dropout} & \textbf{GFlop} & \textbf{CIFAR-10} & \textbf{CIFAR-100} \\ \midrule
Resnext \cite{xie2017aggregated} & 29 & - & - &  & 10.7 & 3.58 & 17.31 \\ \midrule
\multirow{2}{*}{Densenet \cite{huang2017densely}} & 250 & - & - &  & 5.4 & 3.62 & 17.60 \\
 & 190 & - & - &  & 9.3 & 3.46 & \textcolor{blue}{17.18} \\ \midrule
\multirow{3}{*}{WRN \cite{Zagoruyko2016WRN}} & 28 & - & - &  & 5.2 & 4 & 19.25 \\
 & 28 & - & - & \checkmark & 5.2 & 3.89 & 18.85 \\
 & 40 & - & - & \checkmark & 8.1 & 3.8 & 18.3 \\ \midrule
\multirow{3}{*}{WRN-att2 \cite{jetley2018learn}}  & 28 & 2 & - &  & 5.7 & 4.10 & 21.20 \\ 
& 28 & 2 & - & \checkmark & 5.7 & \textbf{3.60} & 20.00 \\
 & 40 & 2 & - & \checkmark & 8.6 & 3.90 & 19.20 \\ \midrule
\multirow{4}{*}{WARN} & 28 & 2 & 4 &  & 5.2 & \textbf{3.60} & 18.72 \\
 & 28 & 3 & 4 &  & 5.3 & \textcolor{blue}{3.45} & 18.61 \\
 & 28 & 3 & 4 & \checkmark & 5.3 & \textcolor{blue}{3.44} & 18.26 \\
 & 40 & 3 & 4 & \checkmark & 8.2 & \textbf{3.46}  & \textbf{17.82}  \\ \bottomrule
\end{tabular}
\end{table}

As seen in Table \ref{tab:accuracy-benchmark}, the proposed Wide Attentional Residual Network (WARN) improves the baseline model for CIFAR-10 and CIFAR-100 even without the use of Dropout and outperforms the rest of the state of the art in CIFAR-10 while being remarkably faster, as it can be seen in Table \ref{tab:performance_benchmark}. Remarkably, the performance on CIFAR-100 makes WARN competitive when compared with Densenet and Resnext, while being up to 36 times faster. We hypothesize that the increase in accuracy of the augmented model is limited by the base network and even better results could be obtained when applied on the best performing baseline. 

\begin{table}[t!]
\centering
\caption{Number of parameters, floating point operations (Flop), time (s) per validation epoch, and error rates (\%) on CIFAR-10 and CIFAR-100. The "Time" column shows the amount of seconds to forward the validation dataset with batch size 256 on a single GPU}
\label{tab:performance_benchmark}
\begin{tabular}{@{}lcccccc@{}}
\toprule
 & \textbf{Depth} & \textbf{Params} & \textbf{GFlop} & \textbf{Time} & \textbf{CIFAR-10} & \textbf{CIFAR-100} \\ \midrule
ResNext & 29 & 68M & 10.7 & 5.02s & 3.58 & 17.31 \\
Densenet & 190 & \textbf{26M} & 9.3 & 6.41s & 3.46 & \textbf{17.18}\\
WRN & 40 & 56M & 8.1 & 0.18s & 3.80 & 18.30\\ 
WRN-att2 & 40 & 64M & 8.6 & 0.24s & 3.90 & 19.20\\ \midrule
WARN & 28 & 37M & \textbf{5.3} & \textbf{0.17s} & \textbf{3.44} & 18.26\\ 
WARN & 40 & 56M & 8.2 & 0.18s & 3.46 & 17.82\\\bottomrule
\end{tabular}
\end{table}

\begin{figure}[!t]
\centering
\begin{subfigure}[!t]{0.45\textwidth}
\centering
\includegraphics[width=0.9\textwidth]{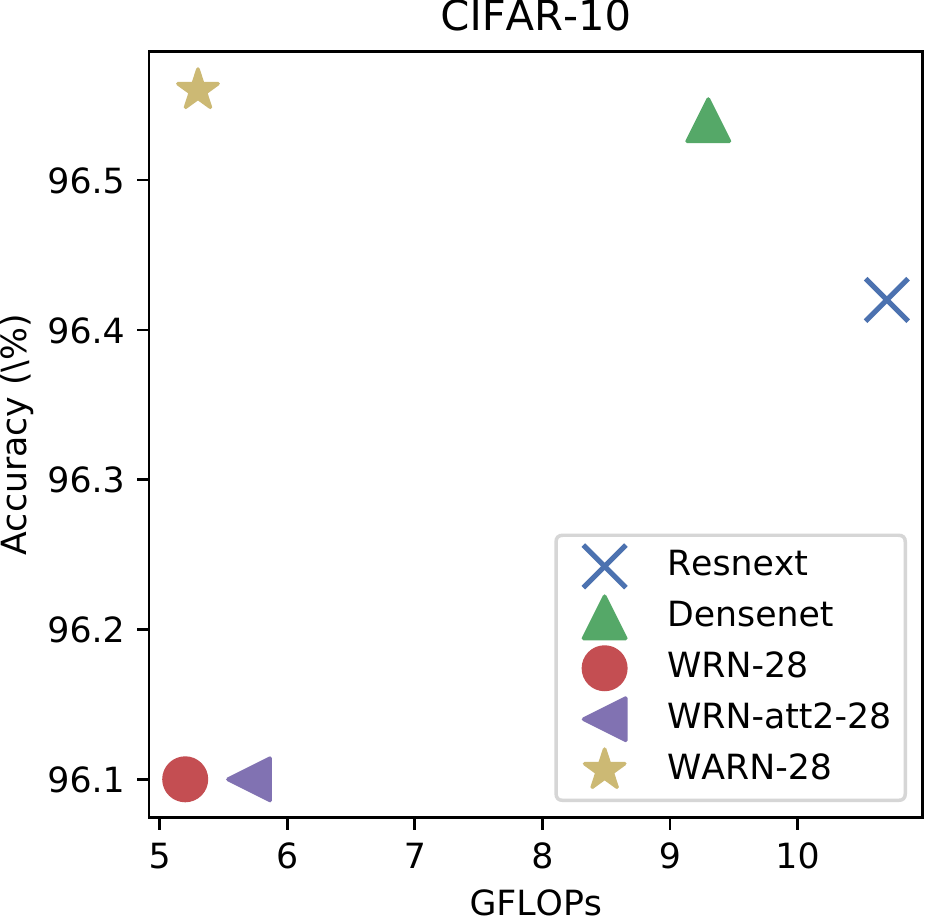}
\caption{CIFAR-10}
\label{fig:clutter_mnist}
\end{subfigure}
\begin{subfigure}[!t]{0.45\textwidth}
\centering
\includegraphics[width=0.9\textwidth]{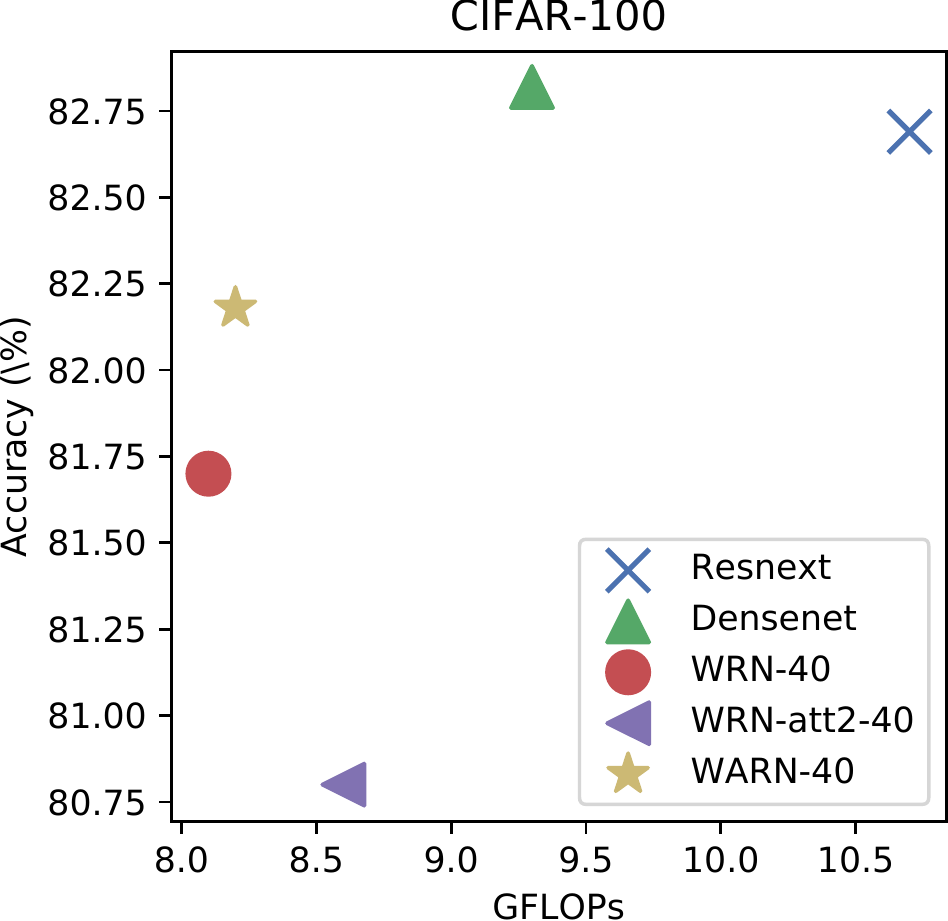}
\caption{CIFAR-100}
\label{fig:clutter_mnist}
\end{subfigure}
\caption{Comparison of the best performing Resnext, Densenet, WRN, WRN-att2, and WARN on the CIFAR-10 and CIFAR-100. Validation accuracy is reported as a function of the number of GFLOPs.}
\label{fig:efficiency}
\end{figure}

Interestingly, WARN shows superior performance even without the use of dropout; this was not possible with \cite{jetley2018learn}, which requires dropout to achieve competitive performances, since they introduce more parameters to the augmented network. The computing efficiency of the top performing models is shown in Figure \ref{fig:efficiency}. WARN provides the highest accuracy per GFlop on CIFAR-10, and is more competitive than WRN, and WRN-att2 on CIFAR-100.

\subsection{Transfer Learning}
\label{sect:transfer}
We fine-tune an augmented WRN-50-4 pre-trained on Imagenet \cite{russakovsky2012imagenet} and report higher accuracy on five different fine-grained datasets: Stanford Dogs, UEC Food-100, Adience, Stanford Cars, CUB200-2001 compared to the WRN baseline. All the experiments are trained for 100 epochs, with a batch size of 64. The learning rate is first set to $10^{-3}$ to all layers except the attention modules and the classifier, for which it ten times higher. The learning rate is reduced by a factor of $0.1$ every 30 iterations and the experiment is automatically stopped if a plateau is reached. The network is trained with standard data augmentation, \emph{i.e.} random $224\times 224$ patches are extracted from $256\times 256$ images with random horizontal flips.

For the sake of clarity and since the aim of this work is to demonstrate that the proposed mechanism universally improves the baseline CNNs for fine-grained recognition, we follow the same training procedure in all datasets. Thus, we do not use $512\times 512$ images which are central for state-of-the-art methods such as RA-CNNs, or MA-CNNs. Accordingly, we do not perform color jitter and other advanced augmentation techniques such as the ones used by \cite{hassannejad2016food} for food recognition. The proposed method is able to obtain state of the art results in Adience Gender, Stanford dogs and UEC Food-100 even when trained with lower resolution.

\begin{table}[t!]
\centering
\caption{Results on six fine-grained recognition tasks. \emph{DSP} means that the cited model uses Domain Specific Pre-training. \emph{HR} means the cited model uses high-resolution images. Accuracies that improve the baseline model are in black bold font, and highest accuracies are in blue}
\label{tab:fine-grained-results}
\begin{tabular}{@{}l|c|c|c|c|c|c@{}}
\toprule
 & \textbf{Dogs} & \textbf{Food} & \textbf{Cars} & \textbf{Gender} & \textbf{Age} & \textbf{Birds} \\ \midrule
\textbf{SotA} & RA-CNN \cite{fu2017look} & Inception \cite{hassannejad2016food} & MA-CNN \cite{zheng2017learning} & FAM \cite{rodriguez2017age} & DEX \cite{Rothe-IJCV-2016} & MA-CNN \cite{zheng2017learning} \\
\textbf{DSP} &  &  &  & \checkmark & \checkmark &  \\
\textbf{HR} & \checkmark &  & \checkmark &  &  & \checkmark \\
\textbf{Accuracy} & 87.3 & 81.5 & \textcolor{blue}{92.8} & 93.0 & \textcolor{blue}{64.0} & \textcolor{blue}{86.5} \\ \midrule
WRN & 89.6 & 84.3 & 88.5 & 93.9 & 57.4 & 84.3 \\
WARN & \textcolor{blue}{92.9} & \textcolor{blue}{85.5} & \textbf{90.0} & \textcolor{blue}{94.6} & \textbf{59.7} & \textbf{85.6} \\ \bottomrule
\end{tabular}
\end{table}

As seen in table \ref{tab:fine-grained-results}, WRN substantially increase their accuracy on all benchmarks by just fine-tuning them with the proposed attention mechanism. Moreover, we report the highest accuracy scores on Stanford Dogs, UEC Food, and Gender recognition, and obtain competitive scores when compared with models that use high resolution images, or domain-specific pre-training. For instance, in \cite{Rothe-IJCV-2016} a domain-specific model pre-trained on millions of faces is used for age recognition, while our baseline is a general-purpose WRN pre-trained on the Imagenet. It is also worth noting that the performance increase on CUB200-2011 ($+1.3\%$) is higher than the one obtained in STNs with $224\times 224$ images ($+0.8\%$) even though we are augmenting a stronger baseline. This points out that the proposed mechanism might be extracting complementary information that is not extracted by the main convolutional stream. As seen in table \ref{tab:acc_param}, WARN not only increases the absolute accuracy, but it provides a high efficiency per introduced parameter.

\begin{table}[t!]
\centering
\caption{Increment of accuracy (\%) per Million of parameters}
\label{tab:acc_param}
\begin{tabular}{@{}lccccccc@{}}
\toprule
\textbf{} & \textbf{Dogs} & \textbf{Food} & \textbf{Cars} & \textbf{Gender} & \textbf{Age} & \textbf{Birds} & \textbf{Average} \\ \midrule
WRN & 1.3 & 1.2 & 1.3 & 1.4 & 0.8 & 1.2 & 1.2 \\
WARN & \textbf{6.9} & \textbf{2.5} & \textbf{3.1} & \textbf{1.5} & \textbf{4.0} & \textbf{2.5} & \textbf{3.4} \\ \bottomrule
\end{tabular}
\end{table}

\begin{figure}[t!]
\centering
\includegraphics[width=\textwidth]{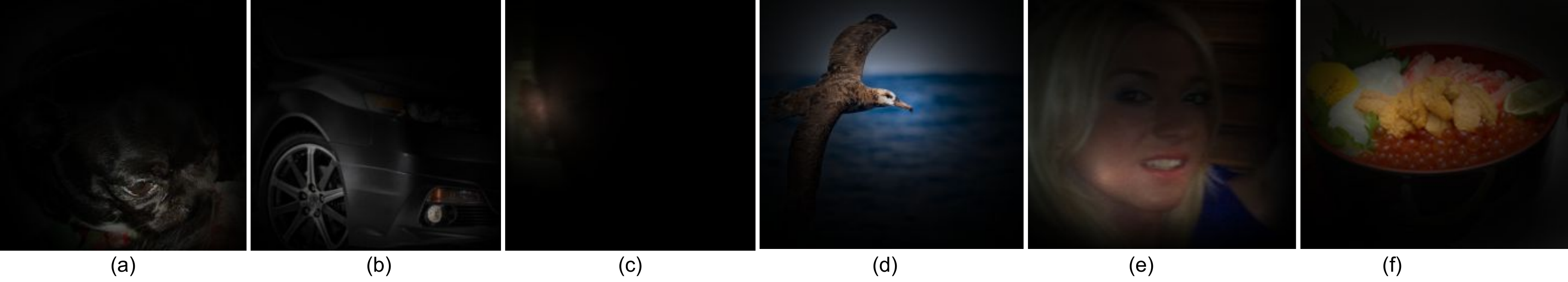}
\caption{Attention masks for each dataset: (a) Stanford dogs, (b) Stanford cars, (c) Adience gender, (d) CUB birds, (e) Adience age, (f) UEC food. As it can be seen, the masks help to focus on the foreground object. In (c), the attention mask focuses on ears for gender recognition, possibly looking for earrings }
\label{fig:attention_masks}
\end{figure}

A sample of the attention masks for each dataset is shown on Figure \ref{fig:attention_masks}. As it can be seen, the attention heads learn to ignore the background and to attend the most discriminative parts of the objects. This matches the conclusions of Section \ref{sect:ablation}.

\section{Conclusion}
We have presented a novel attention mechanism to improve CNNs. The proposed model learns to attend the most informative parts of the CNN feature maps at different depth levels and combines them with a gating function to update the output distribution. Moreover, we generalize attention mechanisms by defining them in two dimensions: the number of attended layers, and the number of Attention Heads per layer and empirically show that classification performance improves by growing the model in these two dimensions. 

We suggest that attention helps to discard noisy uninformative regions, avoiding the network to memorize them. Unlike previous work, the proposed mechanism is modular, architecture independent, fast, simple, and yet WRN augmented with it obtain state-of-the-art results on highly competitive datasets while being 37 times faster than DenseNet, 30 times faster than ResNeXt, and making the augmented model more parameter-efficient. When fine-tuning on a transfer learning task, the attention augmented model showed superior performance in each recognition dataset. Moreover, state of the art performance is obtained on dogs, gender, and food. Results indicate that the model learns to extract local discriminative information that is otherwise lost when traversing the layers of the baseline architecture.

\subsubsection*{Acknowledgments}

Authors acknowledge the support of the Spanish project TIN2015-65464-R (MINECO/FEDER), the 2016FI B 01163 grant of Generalitat de Catalunya, and the COST Action IC1307 iV\&L Net. We also gratefully acknowledge the support of NVIDIA Corporation with the donation of a Tesla K40 GPU and a GTX TITAN GPU, used for this research. 

\bibliographystyle{splncs}
\bibliography{sample.bib}

\end{document}